\documentclass[sigconf]{acmart}
\usepackage{float}
\AtBeginDocument{%
  \providecommand\BibTeX{{%
    \normalfont B\kern-0.5em{\scshape i\kern-0.25em b}\kern-0.8em\TeX}}}

\copyrightyear{2021}
\acmYear{2021}
\setcopyright{acmcopyright}
\acmConference[ADVM '21] {Proceedings of the 1st International Workshop on Adversarial Learning for Multimedia}{October 20, 2021}{Virtual Event, China.}
\acmBooktitle{Proceedings of the 1st International Workshop on Adversarial Learning for Multimedia (ADVM '21), October 20, 2021, Virtual Event, China}
\acmPrice{15.00}
\acmISBN{978-1-4503-8672-2/21/10}
\acmDOI{10.1145/3475724.3483604}

\begin{document}
\fancyhead{}

\title{Imperceptible Adversarial Examples by Spatial Chroma-Shift}

\author{Ayberk Aydin}
\email{aayberk@metu.edu.tr}
\orcid{0000-0002-8083-4536}

\author{Deniz Sen}
\email{deniz.sen_01@metu.edu.tr}
\orcid{0000-0002-5175-3438}

\author{Berat Tuna Karli}
\email{tuna.karli@metu.edu.tr}
\orcid{0000-0003-1697-0477}

\author{Oguz Hanoglu}
\email{oguz.hanoglu@metu.edu.tr}
\orcid{0000-0001-8347-5893}

\author{Alptekin Temizel}
\email{atemizel@metu.edu.tr}
\orcid{0000-0001-6082-2573}
\affiliation{%
  \institution{Graduate School of Informatics, Middle East Technical University}
  \city{Ankara}
  \country{Turkey}
}

\renewcommand{\shortauthors}{Aydin, et al.}

\begin{abstract}

Deep Neural Networks have been shown to be vulnerable to various kinds of adversarial perturbations. In addition to widely studied additive noise based perturbations, adversarial examples can also be created by applying a per pixel spatial drift on input images. While spatial transformation based adversarial examples look more natural to human observers due to absence of additive noise, they still possess visible distortions caused by spatial transformations. Since the human vision is more sensitive to the distortions in the luminance compared to those in chrominance channels, which is one of the main ideas behind the lossy visual multimedia compression standards, we propose a spatial transformation based perturbation method to create adversarial examples by only modifying the color components of an input image. While having competitive fooling rates on CIFAR-10 and NIPS2017 Adversarial Learning Challenge datasets, examples created with the proposed method have better scores with regards to various perceptual quality metrics. Human visual perception studies validate that the examples are more natural looking and often indistinguishable from their original counterparts. 

\end{abstract}



\begin{teaserfigure}
  \includegraphics[width=\textwidth]{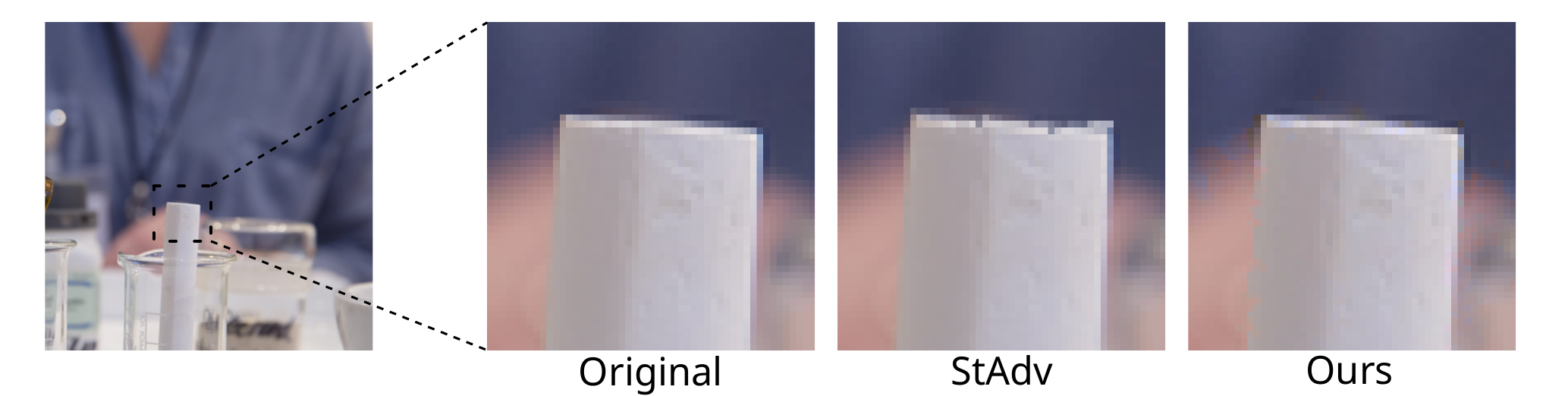}
  \caption{Comparison between the original image and adversarial images obtained by spatially transformations (StAdv) \cite{xiao2018spatially} and spatial chroma-shift (Ours). The proposed method significantly reduces the perceptible deformation while successfully fooling the target models.}
  \label{fig:teaser}
\end{teaserfigure}

\maketitle

\section{Introduction}
Deep Neural Networks have shown impressive performance on several domains such as image classification \cite{he2016resnet}, object detection \cite{ren2015faster}, speech recognition \cite{amodei2016deepspeech} and natural language understanding \cite{devlin2018bert}. However, they are vulnerable to intentionally crafted perturbations called adversarial examples \cite{szegedy2014intriguing} and various methods have been proposed to generate such examples, such as Fast Gradient Sign Method (FGSM) \cite{goodfellow2014explaining}, Carlini\&Wagner Attack \cite{carlini2017towards} and DeepFool \cite{moosavi2016deepfool}. 

Although these examples range from adding a crafted noise to the input to changing the overall image brightness, hue or structure by applying a specified function \cite{laidlaw2019functional}, a common approach is to use $L_p$ norm based metrics to limit the perturbation and impose imperceptibility. Although the use of $L_p$ based norms are debatable and neither necessary nor sufficient criteria for human perception \cite{sharif2018lp}, they are still effective as long as the examples are kept in small $L_p$-norm distances to the original images in the pixel space. However, this brings significant restrictions on perturbations that are not crafted in pixel space. Specifically, spatial transformations, even the most simple ones like translation or rotation, result in high $L_p$ values, while the resulting distortions are hardly perceivable by humans.

Xiao et al. have proposed an alternative approach to perturbation in pixel space, Spatially Transformed Adversarial Examples \cite{xiao2018spatially}, to generate adversarial examples. This approach is based on altering the pixel positions. This is done by applying a flow field regularized by Total Variation based smoothness term, followed by differentiable bilinear interpolation. The prospect that deep  networks can be fooled only by pixel shifts in the image has opened a new research path in generation of adversarial examples. The vulnerability of neural networks to such flow field based spatial transformations raises a natural research question: How far the spatially transformed examples can be optimized for better human perception?

\begin{figure}
\centering
    \begin{tabular}{c c c c}
        \textbf{{Benign}} & \textbf{{UV Only}} & \textbf{{Y Only}} & \textbf{{YUV}}\\
        \includegraphics[scale=0.1]{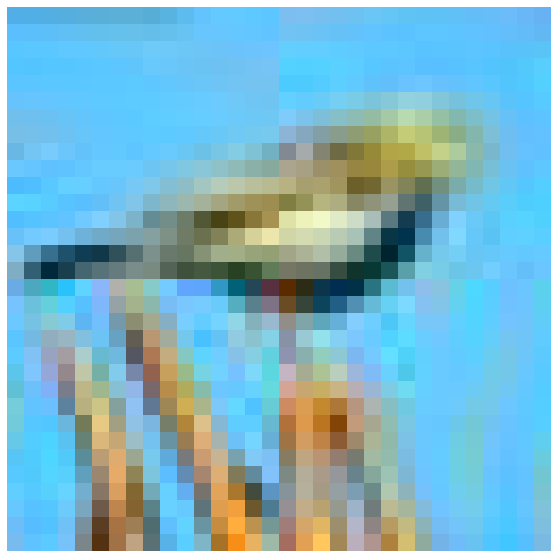} & \includegraphics[scale=0.1]{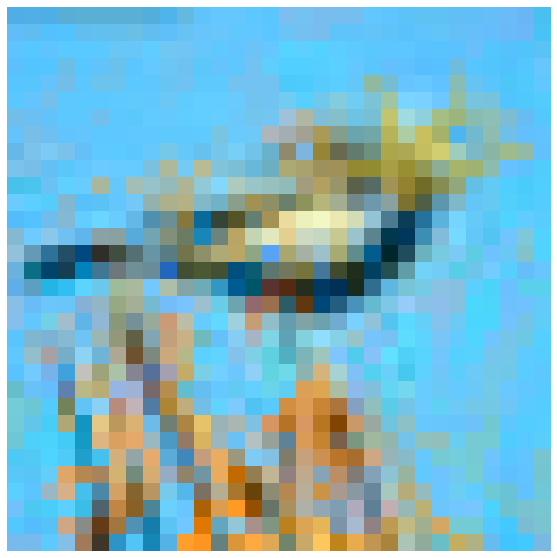} &
        \includegraphics[scale=0.1]{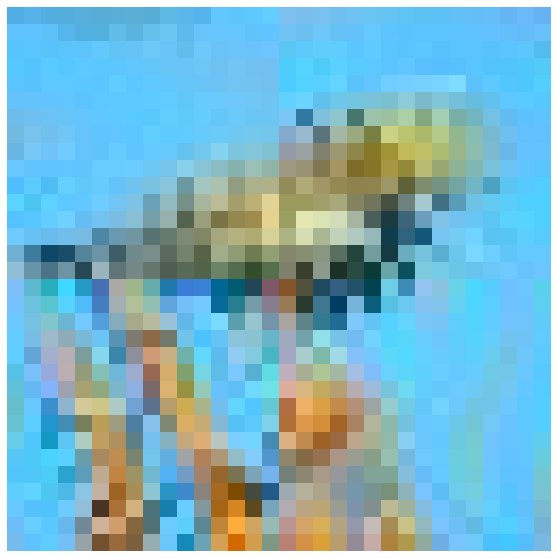} & \includegraphics[scale=0.1]{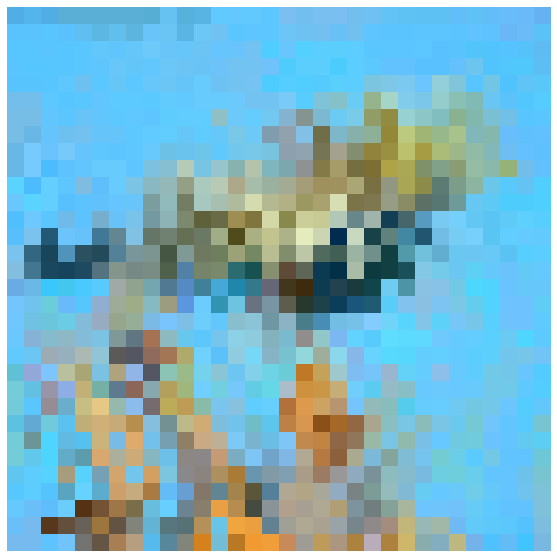}
    \end{tabular}
\caption{Visual comparison of spatially transformed adversarial examples applied on different channels using the identical parameters with exaggerated perturbations for better illustration of the effects (i.e., higher than necessary to fool the model).}
\label{fig:cifar_yuv_channels}
\end{figure}

It is known that the human vision is more sensitive to information loss in the luminance than the chrominance \cite{kasson1992colorspace} and the distortions in the pixel brightness would be more noticeable by humans than the distortions in the color. This has been one of the main motivations behind visual media compression standards \cite{mitchell1992digital} where the images are first converted into YUV domain, where Y channel is the grayscale image (luminance) and U and V channels (chrominance) hold the color information, and, taking human vision into account, a more lossy compression is applied to the U and V components to reduce the data size without compromising perceptual quality. Sample images for a  visual comparison of luminance channel distortions and chroma channel distortions are provided in Figure \ref{fig:cifar_yuv_channels}. Our method succeeded to generate spatially transformed adversarial examples while not producing geometric distortions.

We hypothesize that spatial transformations should not alter the luminance of the input image pixels as a precondition and the grayscale image data should be kept intact. On the other hand, while suppressing the grayscale perturbation increases perceptual quality of generated examples, the unnatural colors and color transitions in adjacent pixels still generates examples which are unnaturally looking and distorted \cite{aksoy2019attack}. 

Motivated by these ideas, we propose a method that  perturbs the input image by spatial transformations only in the colorspace. We show that the method is still able to successfully create adversarial examples both in CIFAR-10 and NIPS2017 Targeted Adversarial Attack Challenge datasets with competitive fooling rates under white-box settings while creating much less visible distortion in comparison to both pixel-value based and spatially transformed examples. In addition, the method alleviates the need for smoothness or \(L_{p}\) norm constraints.

\section{Related work}
\subsection{Adversarial Attacks}
The existence of adversarial examples has been formally demonstrated by Szegedy et al. along with a constrained optimization-based LBFGS attack \cite{szegedy2014intriguing}. Following that work, the widely used single-step adversarial example generation algorithm Fast Gradient Sign Method (FGSM) \cite{goodfellow2014explaining} was introduced, which was further improved by adding several regularization and optimization methods while remaining a one-step attack \cite{tramer2020ensemble, rozsa2016adversarial}. Various iterative versions of FGSM were also introduced, which applied the same gradient ascent optimization in multiple steps \cite{kurakin2017adversarial, dong2018boosting, madry2019pgd}. Another widely used attack is DeepFool, which aims to find the smallest perturbation by carrying the benign example to the closest decision boundary \cite{moosavi2016deepfool}. This algorithm was then modified to learn a single perturbation that can force misclassification on any input image \cite{moosavidezfooli2017universal}. Carlini et al. introduced Carlini\&Wagner (C\&W) attack, which reformulates the constrained optimization problem defined on LBFGS attack \cite{szegedy2014intriguing, carlini2017towards}.

\subsection{Color-based Adversarial Attacks}
Hosseini et al. proposed shifting the hue and saturation components of the HSV color space representation of an input image \cite{hosseini2018semantic}. Another approach is to optimize the perturbations such that the perceptual color distance, which is reported to be a good representation of the human color difference perception. For that, the $L_p$ norm term of the C\&W objective function is replaced by CIEDE2000 perceptual color distance \cite{luo2001development, zhao2020large}. cAdv attack works in CIELAB color space and clusters the image in AB channel representation, then perturbs the clusters whose entropy values are relatively high \cite{bhattad2020unrestricted}. Similarly, ColorFool uses image semantic segmentation to find the color sensitive regions of an image, and shifts the pixel values in the respective LAB color space representation \cite{shamsabadi2020colorfool}. On the other hand, ReColorAdv aims to learn a color transformation function that minimizes the adversarial loss, and applies the same function to each pixel of the input image \cite{laidlaw2019functional}. Aksoy et al. proposed an algorithm in YUV space, where the noise in UV channels are iteratively suppressed. Then Gaussian filters are applied to all the channels to further reduce the $L_2$ distance between the benign and adversarial images \cite{aksoy2019attack}.

\subsection{Transformation-based Adversarial Examples}
While the most common adversarial attacks are done in the pixel domain, i.e. by directly adding noise to individual pixels, deep classifiers have also been shown to have vulnerabilities against spatial deformations and transformations \cite{engstrom2018rotation}. Xiao et al. proposed an algorithm that learns a smooth flow field for the pixel values to shift and produces adversarial examples with less disturbing noise \cite{xiao2018spatially}. Similarly, ADef attack applies spatial perturbations to the image in an iterative fashion \cite{alaifari2019adef}. Zhao et al. proposed an algorithm that takes advantage of both spatial and pixel space distortions \cite{zhao2019perturbations}.

\begin{figure*}
    \centering
    \includegraphics[width=\textwidth,height=\textheight,keepaspectratio]{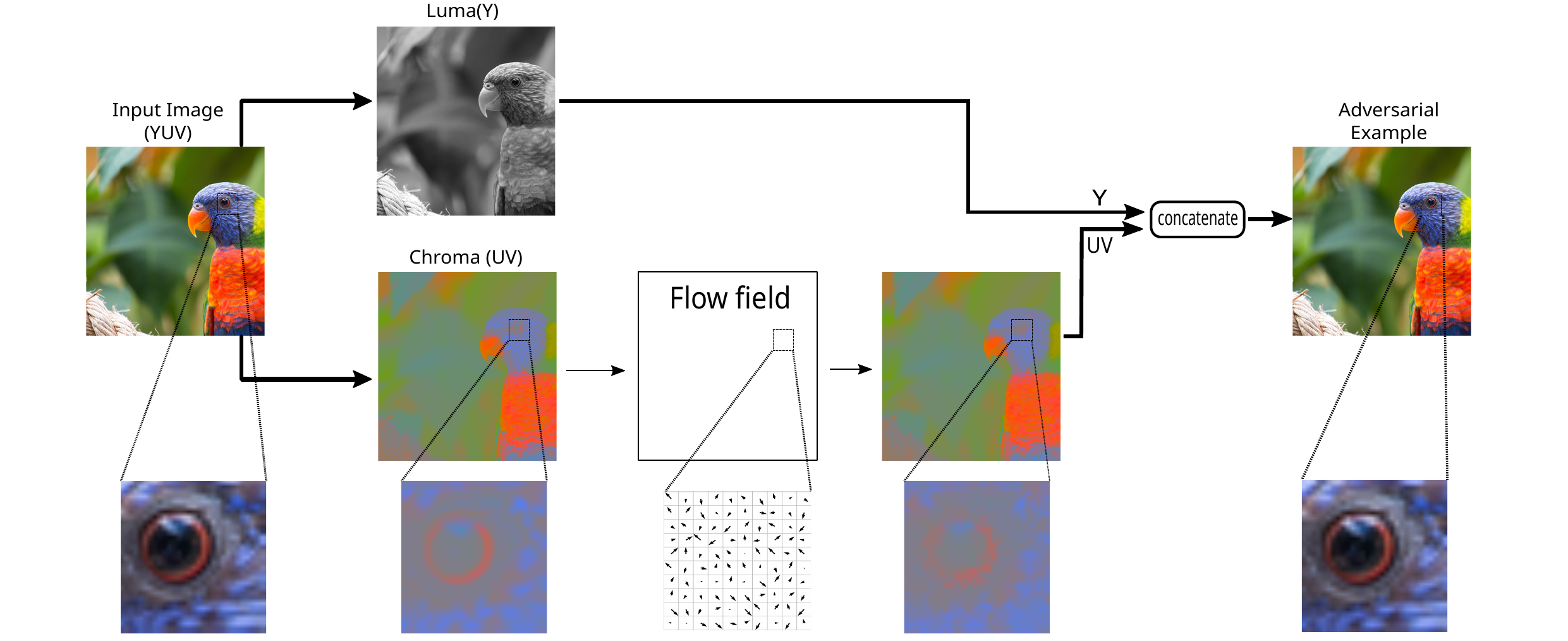}
    \caption{Visual description of the proposed method.}
    \label{fig:visual_algorithm}
\end{figure*}

\section{Methodology}
\subsection{Flow field Calculation}
A flow field \(f\) consists of a two dimensional vector \(f^{(i,j)} = \Delta i,\Delta j\) for each pixel location \((i, j)\) in the adversarial image \(\mathbf{x}_{adv}\). For each pixel in the adversarial image \(\mathbf{x}^{(i,j)}\), a pixel from location \((i + \Delta i, j + \Delta j)\) is sampled from the input image. Since the sampled location is not an integer value, 4 neighboring pixels around the location \((i + \Delta i, j + \Delta j)\) are (bilinear) interpolated \cite{zhou2016view} using Eq. \ref{eq:bilinear} where \(N(i + \Delta i, j + \Delta j)\) is the integer pixel positions around \((i + \Delta i, j + \Delta j)\).

\begin{equation}
\label{eq:bilinear}
\mathbf{x}^{(i,j)}_{adv} = \sum_{\hat{i}, \hat{j} \in N(i + \Delta i, j + \Delta j)} \mathbf{x}^{(\hat{i},\hat{j})} (1 - |\hat{i} - (i + \Delta i)|) (1 - |\hat{j} - (j + \Delta j)|)
\end{equation}

\subsection{Spatial chroma shift}
We aim to generate adversarial examples having little or no visible distortions. To accomplish this, we first convert the input image into YUV representation \(x^{Y,U,V}\) according to BT.470 System M \cite{jack2011video}, with the linear transformation given in Eq. \ref{eq:bilinear}.

\begin{equation}
\label{eq:YUV}
 \left[\begin{array}{c}
Y \\
U \\
V
\end{array}\right]=\left[\begin{array}{ccc}
0.299 & 0.587 & 0.114 \\
-0.14713 & -0.28886 & 0.436 \\
0.615 & -0.51499 & -0.10001
\end{array}\right]\left[\begin{array}{c}
R \\
G \\
B
\end{array}\right]
\end{equation}

Then, we apply a flow field \(f\) to \(x^{U,V}\) to obtain \(x_{adv}^{U,V}\) while keeping \(x^{Y}\) intact. After applying the flow field to the chrominance channels, we concatenate the input image luminance \(x^{Y}\) and adversarial spatially transformed chrominance \(x_{adv}^{U,V}\) channels to obtain the adversarial example \(x_{adv}^{Y,U,V}\). The method is illustrated in Figure \ref{fig:visual_algorithm}. Since the process is end-to-end differentiable, the flow field \(f\) can be optimized by gradient based optimizers. Application of flow field is as explained in \cite{xiao2018spatially}. While a local smoothness constraint is applied in \cite{xiao2018spatially} to enforce adjacent pixels to move in coherence, our method alleviates the need for such constraint and eliminates the need for optimization of this constraint. It also eliminates the need for \(L_{p}\) norm constraints, commonly used in pixel-based perturbation methods, to limit the perturbation.

In addition, the search space can be further reduced to apply subpixel color changes by parameterizing \cite{mordvintsev2018differentiable} the applied flow field \(f\) as in Eq. \ref{eq:tanhf}
\begin{equation}
\label{eq:tanhf}
f = tanh(f')
\end{equation}

and optimizing instead, \(f'\) to minimize adversarial loss. This limits the applied flow field vector size to the range \((-1,1)\) which further reduces visual artifacts of adversarial perturbation without adding any regularization term to the optimized loss value, with a minor decrease in fooling rate.

\section{Experimental Evaluation}
In this section we first quantitatively analyze, and comparatively evaluate the proposed method using learning-based perceptual distance metrics. Then we perform a human perceptual evaluation study for a subjective assessment of the generated adversarial examples. We also evaluate our method against existing defense mechanisms, and compare the results with those of widely known adversarial attacks.

\textbf{Datasets.} We used CIFAR-10 for targeted and NIPS2017 Adversarial Learning Challenge datasets for the untargeted attack evaluations. As the proposed method is designed to work on chroma channels, we excluded the images which are not colorful (i.e., which have a colorfulness measure below the threshold of 15) according to the colorfulness measuring technique in \cite{hasler2003measuring}. This resulted in exclusion of 1467 not-colorful images in CIFAR-10 dataset, leaving 8533 images and in exclusion of 97 not-colorful images in NIPS2017 dataset, leaving 903 images.

\begin{table}
  \caption{: Untargeted adversarial attack results on CIFAR10 testset. Perceptual distances calculated among only fooled examples.}
  \begin{tabular}{*{5}{p{2cm}}}
    \toprule
    Attacks&Fooling Rate&LPIPS&DISTS\\
    \midrule
    Ours (Subpixel) & 88.8\% & \bf{0.006} & \bf{0.019}\\
    Ours (Unrestr.) & 98.1\% & 0.009 & 0.023\\
    stAdv & 97.3\% & 0.021 & 0.048\\
    FGSM & 53.1\% & 0.138 & 0.129\\
    PGD & 100\% & 0.118 & 0.117\\
    DeepFool & 100\% & 0.151 & 0.123\\
    C\&W & 100\% & 0.006 & 0.019\\
  \bottomrule
  \label{tab:cifar}
\end{tabular}
\end{table}

\begin{table}
  \caption{: Targeted adversarial attack results on NIPS2017 dataset. Perceptual distances calculated among only fooled examples.}
  \begin{tabular}{*{5}{p{2cm}}}
    \toprule
    Attacks&Fooling Rate&LPIPS&DISTS\\
    \midrule
    Ours (Subpixel) & 86.8\% & \bf{0.008} & \bf{0.008}\\
    Ours (Unrestr.) & 96.1\% & 0.009 & 0.010\\
    stAdv & 98.8\% & 0.021 & 0.040\\
    C\&W & 100\% & 0.013 & 0.009\\
  \bottomrule
  \label{tab:nips}
\end{tabular}
\end{table}

\textbf{Attacks.} We compare the proposed method against several widely known methods: FGSM \cite{goodfellow2014explaining}, Projected Gradient Descent (PGD) attack \cite{madry2019pgd}, Carlini \& Wagner attack \cite{carlini2017towards}, DeepFool \cite{moosavi2016deepfool} and stAdv \cite{xiao2018spatially}. We used Foolbox \cite{rauber2018foolbox} implementations for FGSM, PGD, C\&W, DeepFool attacks and re-implemented the stAdv attack. We made the implementation publicly available at \url{https://github.com/ayberkydn/stadv-torch} alongside the implementation of the proposed method.

\textbf{Models and Parameters.} We used ResNet50 \cite{he2016resnet} for the CIFAR10 test set and Inception-v3 \cite{szegedy2016inception} for NIPS2017 dataset against all attack types. We set the number of iterations to 1000 and color threshold to 15 for all attack types to test different attacks on an identical ground. We used the default parameters for the existing attacks implemented in the Foolbox. For the proposed methods, we set the learning rate as 0.005 for the CIFAR10 testset and 0.01 for the NIPS2017 dataset.

\textbf{Evaluation.} Regarding that $L_p$ norm distances are both insufficient for perceptual similarity and not applicable to spatially transformed adversarial examples \cite{xiao2018spatially, sharif2018lp}, we have tested perceptual loss of our attack with two learning-based perceptual metrics that are applicable to spatially transformed adversarial examples: Learned Perceptual Image Patch Similarity (LPIPS) metric \cite{zhang2018lpips} and Deep Image Structure and Texture Similarity (DISTS) index \cite{ding2020dists}. LPIPS is a technique that measures Euclidean distance of deep representations (i.e. VGG network \cite{simonyan2015vgg}) calibrated by human perception. LPIPS has been already used on spatially transformed adversarial example studies \cite{laidlaw2019functional, jordan2019quantifying}. DISTS is a method that combines texture similarity with structure similarity (i.e., feature maps) using deep networks with the optimization of human perception. We used the implementation of Ding et al. for the both perceptual metrics \cite{ding2021metrics}.

\textbf{Performance Evaluation with Respect to the Objective Metrics.} Table~\ref{tab:cifar} shows that the proposed method in subpixel mode has the lowest LPIPS and DISTS perceptual loss results, 0.006 and 0.009 respectively, in comparison to FGSM, PGD, DeepFool, C\&W and stAdv attacks on CIFAR10 test set and Table~\ref{tab:nips} shows our subpixel attack has the lowest LPIPS and DISTS results on NIPS2017 dataset in comparison to stAdv and C\&W targeted attacks, 0.008 and 0.008 respectively. In addition to the objective evaluation, we also conducted a human perception experiment, since no perceptual distance metric can accurately represent the human visual perception \cite{carlini2017towards}.

\textbf{Human Perceptual Study.} The experiment has been performed for a subjective evaluation, i.e. whether the chroma-shift based adversarial images are, in most cases, indistinguishable from their original counterparts. 



Our claim is that the perturbations in chroma-shift based adversarial images not only keep the image in the perceptually realistic domain but also go beyond this by making the differences almost imperceptible to human eye. We aim to test this claim and compare the results with those of stAdv, which was already shown to keep the images in perceptually realistic domain \cite{xiao2018spatially}. 
In our human perception study, we follow a similar protocol to this work with a minor difference that there is no time limit for the participants and images stay on the screen until the participants make their choices. This allows participants to analyze the images in more detail for any visible distortions.

\begin{figure*}[ht]
\centering
    \begin{tabular}{c c c}
        \textbf{{Benign}} & \textbf{{stAdv}} & \textbf{{Subpixel}}\\
        \includegraphics[scale=0.6]{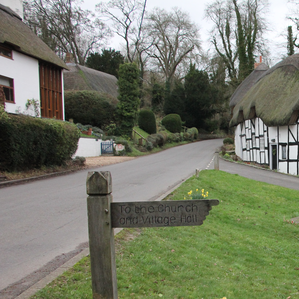} & \includegraphics[scale=0.6]{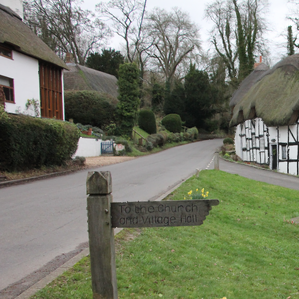} &
        \includegraphics[scale=0.6]{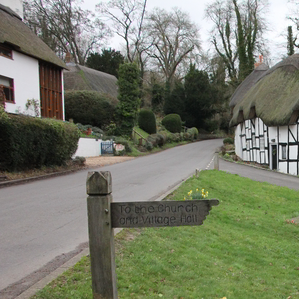}
    \end{tabular}
\caption{A sample of benign, stAdv applied and subpixel chroma-shift applied images from the human perceptual study, taken from NIPS2017 dataset.}
\label{fig:nips}
\end{figure*}

For the experiments, we used the first 10 images from the NIPS 2017 competition dataset. For each original image, we prepared an adversarial images using stAdv and another using the proposed chroma-shift based method, an example image and their adversarial counterparts are shown in Figure \ref{fig:nips}. The participants are presented with two images, one of which is always the original image and the other one shown alongside is either the same image again, stAdv image, or chroma-shifted image in a random order. The participants are asked the question “Is the image on the right the same with the one on the left or a slightly distorted version of it?” for each pair. This results in 30 annotations per participant.

\begin{figure*}[t]
    \centering
    \includegraphics[scale=.45]{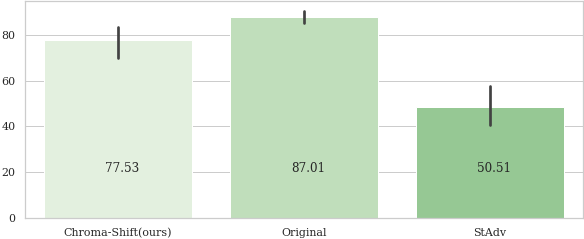}
    \caption{Human perceptual study results where the bars signify the percentage of the participants answering `The images are the same'}
    \label{fig:human_study}
\end{figure*}

A total of 77 participants participated in the study, providing a total of 1860 annotations. As shown in Figure \ref{fig:human_study}, the chroma-shift based adversarial examples are perceived frequently the same as the original images. Note that in some cases the original images (i.e., control group) were perceived differently and 87.01\% of the total annotations the  original vs. original were annotated the same. While this reduced down to 77.53\% for the original vs. chroma-shift images, most were still indistinguishable by the participants. For the original vs. stAdv images, the participants were able to spot the distortion with more ease  and 50.51\% of the total annotations stated that they were the same.

\textbf{Performance Against Robust Models.}
We also measured the performance of our attack against adversarially trained ResNet50 models. To test the robustness of our attack against adversarially trained models, we have used 3 robust models that are trained with FGSM, 20-step PGD and ensemble adversarial training methods. The models were attacked in an $L_\infty$ setting where $\epsilon$ budget is set to be 8/255 (in 0-1 scale). Ensemble adversarial training \cite{tramer2020ensemble} was performed on ResNet50, where the generator models were set to be DenseNet121 \cite{huang2018densenet}, GoogLeNet \cite{szegedy2015going} and MobileNet \cite{howard2017mobilenets}. Table~\ref{tab:robust} shows the fooling rates obtained by attacking these robust models using different adversarial attacks. Chroma-shift attack, when run in unrestricted mode, has on par or better than C\&W attack, it also outperformed stAdv attack against each augmentation based adversarial defense mechanism. 

\begin{table}[h]
    \centering
        \caption{Fooling rates of different types of adversarial attacks on adversarially trained ResNet50 classifiers}

        \begin{tabular}{*{6}{p{1cm}}}
        \toprule 
        \text {Defense} & \text {FGSM} & \text {C\&W} & \text {stAdv} & \text {Subpixel} & \text {Unrestr.} \\
        \midrule
        \text {FGSM}     & 57\% & 66\% & 30\% & 45\% & \bf{ 73\%} \\
        \text {PGD}      & 32\% & \bf{68\%} & 27\% & 67\% & 67\% \\
        \text {Ensemble} & 92\% & 91\% & 74\% & 72\% & \bf{95\%} \\
        \bottomrule 
        \end{tabular}
    \label{tab:robust}
\end{table}

\textbf{Grayscale defense.} Adversarial examples applied only on image colors (i.e., preserving the luminance content and modifying only the chroma channels) are expected to be not robust to conversion to grayscale as mentioned in the study of Laidlaw et al. \cite{laidlaw2019functional}. As the proposed method is applied to only chrominance (UV) channels, it could be argued that converting the input images from RGB to grayscale could be used as a defence. However, conversion to grayscale may be counterproductive as it has an undesired effect of decreasing the overall accuracy on both benign and adversarial examples. When grayscale versions of the input images are used, CIFAR10 testset accuracy decreases down to 86.3\% from 93.7\% and NIPS2017 dataset accuracy decreases to 79.5\% from 92.6\%.

\section{Discussions}

Our hypothesis was that it is possible to create adversarial examples that contain visually imperceptible perturbations by only shifting the chrominance values of the original image. The results obtained with targeted and untargeted attacks show that our hypothesis is valid, there is a decrease in the perceptual distances with regards to the LPIPS and DISTS metrics with competitive fooling rates. Furthermore, the common expectation would be in the way that 'adversarial energy' are mostly accumulated in the luminance channel, which is where the perturbations naturally appear when adversarial attacks are applied in the RGB space\cite{pestana2020adversarial}. While this assumption is still valid, our findings suggest that chrominance channels contain sufficient adversarial energy so that the network can be fooled by only perturbing the UV channels. On the other hand, our attack is still disadvantageous in fooling rate compared to C\&W attack, which is a pixel space attack whereas our algorithm does optimization in the vector space. 

We also validated our hypothesis by performing a human visual study, as despite both perceptual metrics we have used have been shown to be correlated with human vision, no perceptual metric is a precise measure for human vision \cite{carlini2017towards}. Results shown in Figure \ref{fig:human_study}, signify that our algorithm not only makes improvement on quantitative visual metrics, but also performs notably well in practice and it is robust against widely used adversarial training based defense mechanisms.

\section{Conclusions}
In this work, we have shown that small spatial transformations exclusively in the color space can yield adversarial examples that are often indistinguishable from their original counterparts. We compared the proposed attack with widely used attacks, both for standard and adversarially trained models, using different perceptual difference distances. Furthermore, we have evaluated the visual quality of our adversarial examples with a subjective human experiment; which showed that chroma-shifted adversarial examples are indistinguishable from the original images in most cases. While conversion to grayscale at the input could be used as a defence against the proposed attack, it has to be noted that this also results in an overall decrease in the standard accuracy. In the future, to further weaken such defense, the attack could be combined with traditional $L_p$ norm attacks such as the method proposed in \cite{jordan2019quantifying}. As another alternative, some limited amounts of perturbations in Y channel could be included in addition to the perturbations in UV channels, similar to the implementation of Aksoy et al. \cite{aksoy2019attack}.


\begin{acks}
This work has been funded by The Scientific and Technological Research Council of Turkey, ARDEB 1001 Research Projects Programme project no: 120E093
\end{acks}

\bibliographystyle{ACM-Reference-Format}
\bibliography{referanslar}

\end{document}